\documentclass[10pt,twocolumn,letterpaper]{article}
\usepackage[OT1]{fontenc}
\usepackage[pagenumbers]{cvpr}
\usepackage{amsmath}
\usepackage{amssymb}
\usepackage{booktabs}
\usepackage{multirow}
\usepackage{float}
\usepackage{placeins}
\usepackage{graphicx}
\usepackage{tikz}
\usetikzlibrary{arrows.meta,calc,fit,positioning}
\usepackage{xcolor}
\usepackage{xspace}
\usepackage[table]{xcolor}

\definecolor{radarAccuracy}{RGB}{210,78,78}
\definecolor{radarSpeed}{RGB}{70,96,170}
\definecolor{radarAccuracyEdge}{RGB}{170,38,48}
\definecolor{radarSpeedEdge}{RGB}{42,58,128}
\definecolor{radarTokens}{RGB}{73,145,118}
\definecolor{radarTokensEdge}{RGB}{32,108,81}
\definecolor{radarGroup}{RGB}{238,174,105}

\definecolor{riseBlue}{RGB}{62,105,176}
\definecolor{riseViolet}{RGB}{118,91,171}
\definecolor{riseGreen}{RGB}{47,137,109}
\definecolor{riseOrange}{RGB}{216,137,66}
\definecolor{riseInk}{RGB}{50,58,72}

\definecolor{cvprblue}{rgb}{0.21,0.49,0.74}
\usepackage[pagebackref,breaklinks,colorlinks,allcolors=cvprblue]{hyperref}
\def\confName{CVPR}
\def\confYear{2027}
\title{RISE: Reasoning with Implicit State Embeddings for GUI Agents}
\hypersetup{pdftitle={RISE: Reasoning with Implicit State Embeddings for GUI Agents},pdfauthor={Anonymous manuscript}}

\author{
Enqi Liu$^{1,2}$, 
Zhi Gao$^{1,2}$\footnotemark[1], 
Liu Liu$^{4}$,
Lifeng Fan$^{2}$, 
Liyuan Pan$^{1,3}$\thanks{Corresponding author.}, \\
$^{1}$ Beijing Institute of Technology, Beijing, China\\
$^{2}$ Beijing Institute for General Artificial Intelligence, Beijing, China\\
$^{3}$ Yangtze Delta Region Academy of Beijing Institute of Technology, Jiaxing, China \\
$^{4}$ Huawei Technologies Ltd, Beijing, China\\
{\tt\small \{enqi.liu, zhi.gao, liyuan.pan\}@bit.edu, liuliu33@huawei.com, lifengfan@bigai.ai}
}

\usepackage{needspace}
\usepackage{ragged2e}
\begin{document}
% Keep paragraph spacing natural from the first page.
\raggedbottom
\maketitle
\begin{abstract}
Graphical user interface (GUI) agents translate user instructions and screenshots into executable actions. To learn executable actions, existing approaches often rely on explicit reasoning during training, which encourage over-reliance on specific reasoning patterns rather than generalizable action-relevant information. In this paper, we propose an implicit framework, RISE, which integrates action-relevant reasoning into implicit state representations to guide GUI action prediction. To this end, action-relevant reasoning is captured through three complementary supervision objectives: full-CoT state alignment preserves information across the reasoning process, action-distribution alignment encodes action preferences, and reasoning-to-action boundary alignment captures the representation immediately before action generation. In this case, our RISE effectively learns informative implicit representations, thereby improving the accuracy of next-action token prediction. RISE achieves 70.6\% accuracy on OSWorld-G, establishing state-of-the-art performance among compared 8B models with a 4.3-point gain over its backbone. Results across two distinct backbones and six diverse benchmarks further demonstrate its effectiveness in GUI grounding and action prediction. Our code is available on \href{https://github.com/Liu-arch/RISE}{our project page}.
\end{abstract}
\begin{figure}[!t]
\centering
\includegraphics[width=\columnwidth]{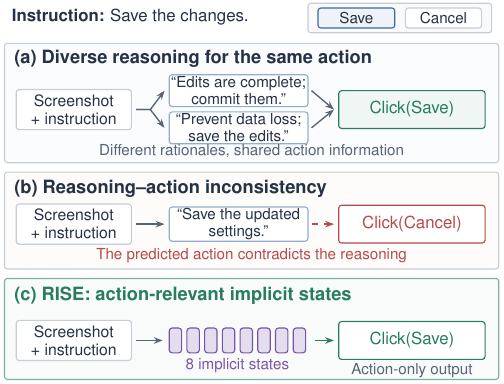}

\caption{\textbf{Illustrative examples} show two challenges: (a) different reasoning expressions can support the same action; (b) generated reasoning can suggest an action that contradicts the final prediction. (c) RISE learns action-relevant implicit states to guide GUI action prediction.}

\label{fig:concept}
\end{figure}
% \section{Introduction} 

\section{Introduction}

With the growing adoption of graphical user interface (GUI) agents across web browsers~\cite{deng2023mind2web,xu2024aguvis}, desktop applications~\cite{xie2024osworld,wang2025opencua}, and mobile devices~\cite{rawles2024androidworld,li2024androidcontrol}, developing accurate and efficient action prediction methods has attracted increasing attention. GUI agents translate screenshots and user instructions into executable actions, which require selecting operation types and predicting their associated arguments, such as target coordinates and text inputs. Existing approaches commonly employ autoregressive vision-language models (VLMs) to generate actions as textual sequences in a reasoning process~\cite{lin2025showui,qin2025uitars,yang2026guilibra}. These methods typically serialize thoughts, operation types, coordinate values, and free-text inputs through a unified token-by-token generation process~\cite{cheng2024seeclick}.

Despite these advances, such an explicit reasoning process does not necessarily ensure reliable alignment between reasoning and GUI action prediction. This limitation manifests in two aspects. 
(1) Reasoning diversity: Explicit reasoning can express the same intended action through different reasoning paths, introducing variability in textual reasoning even when the target action remains unchanged. Learning on explicit reasoning may encourage reliance on specific reasoning expressions rather than shared action-relevant information. As illustrated in Fig.~\ref{fig:concept}(a), the instruction `Save the changes' may lead to different reasoning paths, such as `Edits are complete, commit them' and `Prevent data loss, save the edits', both resulting in the same action, \texttt{Click(Save)}.
(2) Weak reasoning--action coupling: Although actions are generated following explicit reasoning, conventional token-level supervision does not explicitly enforce semantic consistency between reasoning and the predicted action. Consequently, as shown in Fig.~\ref{fig:concept}(b), the generated reasoning `Save the updated settings' suggests saving the changes, yet the model predicts \texttt{Click(Cancel)}, contradicting its preceding reasoning.

In this paper, we explore implicit reasoning for GUI action prediction. Compared with explicit reasoning generation, implicit representations can capture action-relevant information without relying on specific textual expressions. By learning from action prediction objectives, implicit states can encode reasoning information in a form directly useful for action prediction, without requiring intermediate reasoning text. This motivates us to leverage implicit states to guide GUI action prediction.

This objective presents two key challenges. (1) Capturing shared reasoning information: Reasoning paths leading to the same action may differ substantially in length, wording, and intermediate steps, making it difficult to condense their shared action-relevant information into a fixed number of implicit states. These states must also retain the visual and instructional context needed for action prediction. (2) Ensuring effective reasoning--action coupling: Aligning implicit states with explicit reasoning representations alone does not guarantee that the action decoder will utilize the learned information. The implicit states must directly contribute to operation selection and visual target localization without losing the fine-grained visual information required for accurate grounding.

In this paper, we propose RISE (Reasoning with Implicit State Embeddings), a framework that learns action-relevant reasoning through implicit states, as illustrated in Fig.~\ref{fig:concept}(c). Specifically, RISE introduces learnable embeddings that are contextualized by the current screenshot and instruction to construct input-dependent implicit states.
We first employ full-CoT state alignment to condense information across the reasoning process into a fixed number of implicit states, allowing the model to learn from reasoning without reproducing specific textual expressions. We then introduce action-distribution supervision to capture action preferences, encouraging the implicit states to retain information useful for distinguishing among possible actions rather than merely representing the reasoning process. We further perform reasoning-to-action boundary alignment to capture the representation immediately before action generation, providing complementary supervision on how reasoning information is organized for action prediction. Together, these strategies enable RISE to leverage implicit reasoning alongside the original visual and textual information for GUI action prediction without generating intermediate reasoning text.

Our RISE achieves 70.6\% accuracy on OSWorld-G, establishing state-of-the-art performance among the compared 8B models and outperforming its backbone baseline by 4.3 percentage points. Extensive experiments across six benchmarks further demonstrate the effectiveness of RISE in GUI grounding and action prediction.

Our main contributions are as follows:
\begin{itemize}
    \item We propose RISE, an implicit reasoning framework that learns action-relevant information through contextualized implicit states to guide GUI action prediction without generating intermediate reasoning text.
    \item We introduce three complementary supervision strategies that connect intermediate reasoning representations, action distributions, and reasoning-to-action boundary states to action prediction.
    \item We demonstrate improvements in GUI grounding and offline action prediction across two backbones and six benchmarks, supported by matched comparisons, supervision ablations, and inference-efficiency analyses.
\end{itemize}

\begin{figure*}[t]
\centering
\includegraphics[width=0.98\textwidth]{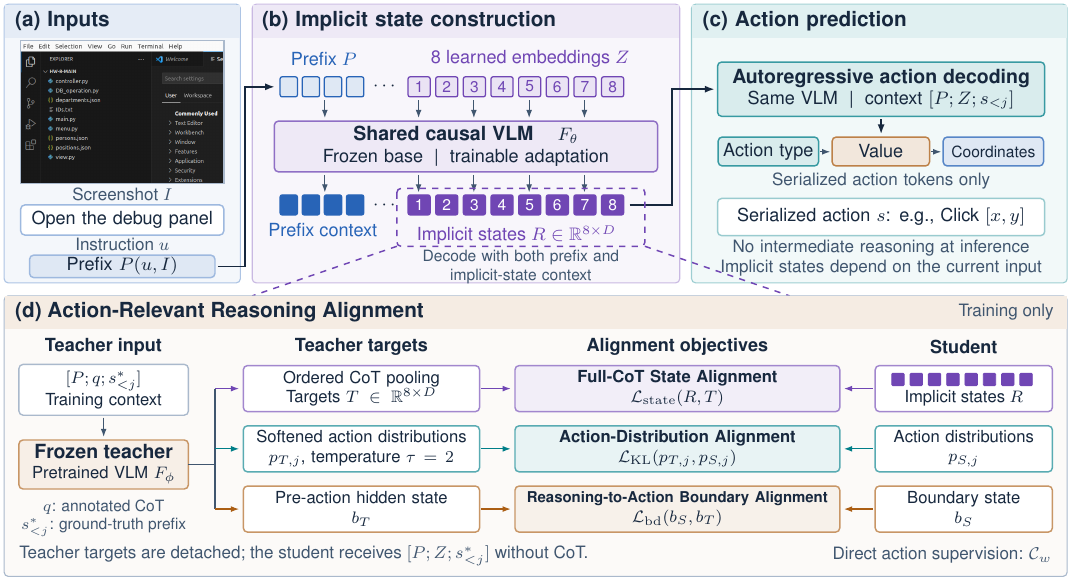}
\caption{\textbf{Overview of RISE.} (a) The screenshot and instruction form the multimodal prefix $P$. (b) The student contextualizes eight learned embeddings $Z$ to obtain implicit states $R$. (c) Action tokens are generated using both the original prefix and implicit-state context. (d) Action-Relevant Reasoning Alignment uses a frozen copy of the pretrained VLM as the teacher. Its input contains $P$, annotated CoT $q$, and preceding ground-truth action tokens $s_{<j}^*$. Teacher targets are aligned with the student's implicit states, action-token distributions, and pre-action boundary state through $\mathcal L_{\mathrm{state}}$, $\mathcal L_{\mathrm{KL}}$, and $\mathcal L_{\mathrm{bd}}$, respectively. 
% Action cross-entropy provides direct supervision for action prediction. 
% At inference, only the student is used, without CoT input or intermediate reasoning generation.
}
\label{fig:latent}
\end{figure*}

\section{Related Work}
\par\noindent\textbf{GUI Action Learning.}\quad
Visual GUI agents learn to map screenshots and user instructions to executable actions, including operation selection and target localization. Early studies establish visual grounding capabilities through screenshot--language pretraining and cross-platform interaction data~\cite{hong2023cogagent,cheng2024seeclick,gou2024uground,wu2024osatlas}. Recent approaches further incorporate task interpretation and intermediate reasoning into action demonstrations to improve GUI action prediction~\cite{xu2024aguvis,qin2025uitars,yang2026guilibra,liu2026benchmarking}. Feedback-based methods enhance action learning through preference optimization or reward-guided training~\cite{qin2025uitars,lu2025uir1,luo2025guir1,liu2025infiguir1}. While these studies advance GUI action learning through richer demonstrations and optimization strategies, RISE focuses on how action-relevant reasoning information is represented and utilized. Specifically, it learns implicit reasoning states that guide action prediction alongside the original screenshot and instruction, without generating intermediate reasoning text.

\par\noindent\textbf{Implicit Reasoning.}\quad
Implicit reasoning models intermediate computation through continuous representations rather than explicit reasoning text. Prior studies explore learning implicit representations through hidden-state supervision and continuous CoT compression~\cite{deng2023implicitcot,shen2025codi,shen2025heima}, as well as recurrent and parallel latent reasoning mechanisms~\cite{hao2024coconut,wu2025pccot}. Learnable continuous embeddings and additional computation positions provide alternative ways to condition model predictions~\cite{li2021prefixtuning,lester2021prompttuning,goyal2024pause}. In GUI interaction, MIRAGE combines latent reasoning with a generative world-model objective that aligns reasoning representations with future screenshots~\cite{yang2026mirage}. Different from these approaches, RISE learns screenshot- and instruction-conditioned implicit states using complementary supervision from the full reasoning process, action distributions, and reasoning-to-action boundary states, enabling reasoning information to guide GUI action prediction without generating intermediate reasoning text.

\section{Method}
\label{sec:method}
An overview of RISE is presented in Fig.~\ref{fig:latent}. We first formulate GUI action prediction, then describe the construction of implicit states and three complementary supervision strategies, and finally present the training objective.

\subsection{Formulation}
\label{sec:problem-formulation}

Given a screenshot, a user instruction, and any available interaction history, the GUI agent constructs a multimodal prefix $P\in\mathbb{R}^{n\times D}$, where $n$ is the input sequence length and $D$ is the hidden dimension. Let $s=(s_1,\ldots,s_L)$ denote an executable action serialized in the backbone's native output format, where $s_j$ is the token at position $j$ and $L$ is the sequence length.

RISE conditions action prediction on the multimodal prefix $P$ and $K$ learnable embeddings $Z\in\mathbb{R}^{K\times D}$. The action-generation policy factorizes autoregressively as
\begin{equation}
\pi_\theta(s\mid P,Z) = \prod_{j=1}^{L} \pi_\theta(s_j\mid P,Z,s_{<j}),
\label{eq:autoregressive-action}
\end{equation}
where $s_{<j}$ contains the preceding action tokens.

\subsection{Implicit State Construction}
\label{sec:implicit-interface}

RISE appends the learnable embeddings $Z$ to the multimodal prefix $P$ to construct contextualized implicit states:
\begin{equation}
R=F_\theta([P;Z])_{n+1:n+K},
\label{eq:latent-interface}
\end{equation}
where $F_\theta$ denotes the VLM backbone, $[P;Z]$ denotes sequence concatenation, and the subscript selects the $K$ appended hidden states. The resulting $R\in\mathbb{R}^{K\times D}$ contains the contextual implicit states. 
% While $Z$ is shared across examples and learned during training, $R$ is computed anew for each input.

Through causal attention, each action token can attend to both the original multimodal prefix and the implicit states, allowing the model to leverage implicit reasoning alongside the original visual and textual information. By default, we use $K=8$ implicit positions.

\subsection{Action-Relevant Reasoning Alignment}
\label{sec:fullcot-distillation}

We use a frozen copy of the pretrained VLM as the teacher to provide three complementary supervision targets for learning action-relevant implicit states. Let $q$ denote the annotated CoT associated with each training example and $\phi$ the teacher parameters. Given the multimodal prefix $P$, the teacher processes $[P;q]$, while the RISE student conditions on $P$ and learnable embeddings $Z$ without accessing explicit CoT. 
RISE learns action-relevant implicit states from the teacher's full CoT representations, action-token probability distributions, and reasoning-to-action boundary state.

\par\noindent\textbf{Full-CoT State Alignment.}
We align the student's implicit states $R$ with representations aggregated from the teacher's hidden states across the full CoT. Specifically, we pool the teacher's CoT hidden states into $K$ ordered target representations $T\in\mathbb{R}^{K\times D}$, excluding reasoning delimiters and action tokens. Let $M$ denote the number of teacher hidden states within the CoT body. When $M \geq K$, we average the states within each of $K$ contiguous segments. When $M < K$, we sample $K$ uniformly spaced positions and round them to token indices, reusing states as needed. This pooling preserves the order of the reasoning sequence.

The state-alignment objective is defined as
\begin{equation}
\mathcal{L}_{\mathrm{state}}
=
\frac{1}{K}\sum_{i=1}^{K}
\left(1-\cos(R_i,T_i)\right)
+
0.1\mathcal{H}(R,T),
\label{eq:state-alignment}
\end{equation}
where $\cos(\cdot,\cdot)$ denotes cosine similarity and $\mathcal{H}(\cdot,\cdot)$ denotes the element-wise mean SmoothL1 loss with unit transition. This objective encourages the implicit states to retain information represented throughout the teacher's reasoning without requiring the student to reproduce its wording.

\par\noindent\textbf{Action-Distribution Alignment.}
Beyond aligning implicit states with reasoning representations, we further supervise the action predictions through their probability distributions. Let $\ell_{T,j}$ and $\ell_{S,j}$ denote the teacher and student vocabulary logits at action-token position $j$. 
Both models use the same ground-truth action prefix $s_{<j}^*$ during teacher-forced training.
Their softened distributions are
\begin{equation}
p_{T,j}=\operatorname{softmax}(\ell_{T,j}/\tau),
p_{S,j}=\operatorname{softmax}(\ell_{S,j}/\tau),
\end{equation}
where $\tau=2$ is the temperature used to soften the action distributions. We minimize the teacher-to-student divergence:
\begin{equation}
\mathcal{L}_{\mathrm{KL}}=\frac{\tau^2}{\sum_j w_j}\sum_j w_j\operatorname{KL}(p_{T,j}\Vert p_{S,j}).
\end{equation}
Here the sums cover supervised action-token positions in the training batch. The weight $w_j$ is 2 for tokens wholly within action-type values, 4 for tokens wholly within coordinate values, and 1 otherwise. The $\tau^2$ factor compensates for the temperature-dependent scaling of the gradients~\cite{hinton2015distilling}. 
This objective encourages the implicit states to support accurate action prediction beyond matching intermediate reasoning representations.

\par\noindent\textbf{Reasoning-to-Action Boundary Alignment.}
To capture the transition from reasoning to action generation, we align the teacher's hidden state after processing the annotated CoT with the hidden state of RISE after processing the multimodal prefix and implicit states. Both states are taken immediately before the first action token is predicted. Let $b_T\in\mathbb{R}^{D}$ and $b_S\in\mathbb{R}^{D}$ denote these teacher and student boundary states, respectively. The boundary-alignment objective is
\begin{equation}
\mathcal{L}_{\mathrm{bd}}
=
1-\cos(b_S,b_T)
+
0.1\mathcal{H}(b_S,b_T).
\label{eq:boundary-alignment}
\end{equation}
This objective captures the transition from reasoning to action generation, encouraging the implicit states to encode reasoning information relevant to the final action prediction.

\subsection{Training}
\label{sec:training}

We supervise action prediction using ground-truth actions, allowing the implicit states to learn action-relevant information through the action prediction objective.
% Ground-truth actions provide direct supervision for the implicit states and action prediction. 
For a target action sequence $s^*$, the weighted action cross-entropy is
\begin{equation}
\mathcal{C}_w=-\frac{1}{\sum_j w_j}\sum_j w_j\log\pi_\theta(s_j^*\mid P,Z,s_{<j}^*),
\label{eq:c0}
\end{equation}
where $s_{<j}^*$ contains the preceding ground-truth action tokens. The summation and weights follow the action-distribution objective described above.

We jointly optimize action prediction and the three complementary reasoning supervision objectives:
\begin{equation}
\mathcal{L} = \mathcal{C}_w
+ \lambda_{\mathrm{state}}\mathcal{L}_{\mathrm{state}}
+ \lambda_{\mathrm{KL}}\mathcal{L}_{\mathrm{KL}}
+ \lambda_{\mathrm{bd}}\mathcal{L}_{\mathrm{bd}},
\label{eq:distill-objective}
\end{equation}
where $\lambda_{\mathrm{state}}=0.1$, $\lambda_{\mathrm{KL}}=0.5$, and $\lambda_{\mathrm{bd}}=0.25$ are the corresponding loss weights.

\begin{table*}[t]
\centering\footnotesize
\caption{OSWorld-G (Original) accuracy (\%) by task category and overall.
Task categories are not mutually exclusive; Overall is computed over all 564 examples. Published references retain their original protocols.
Bold scores indicate the best results, while underlined scores indicate the second-best results in each column.}
\label{tab:osworldg-main}
\setlength{\tabcolsep}{2.5pt}
\renewcommand{\arraystretch}{1.08}
\begin{tabular*}{\textwidth}{@{\extracolsep{\fill}}lrrrrrr@{}}
\toprule
\multirow{2}{*}{Method / configuration} & \multicolumn{5}{c}{Task-category accuracy (\%)$\uparrow$} & \multirow{2}{*}{Overall$\uparrow$}\\
\cmidrule(lr){2-6}
& \shortstack{Text\\matching} & \shortstack{Element\\recognition} & \shortstack{Layout\\understanding} & \shortstack{Fine-grained\\manipulation} & Refusal &\\
\midrule
\rowcolor{black!7}
\multicolumn{7}{@{}l}{External baselines}\\
OS-Atlas-7B~\cite{wu2024osatlas} & 44.1 & 29.4 & 35.2 & 16.8 & 7.4 & 27.7\\
Qwen2.5-VL-7B~\cite{bai2025qwen25vl} & 45.6 & 32.7 & 41.9 & 18.1 & 0.0 & 31.4\\
UGround-V1-7B~\cite{gou2024uground} & 51.3 & 40.3 & 43.5 & 24.8 & 0.0 & 36.4\\
Aguvis-7B~\cite{xu2024aguvis} & 55.9 & 41.2 & 43.9 & 28.2 & 0.0 & 38.7\\
UI-TARS-7B~\cite{qin2025uitars} & 60.2 & 51.8 & 54.9 & 35.6 & 0.0 & 47.5\\
GUI-Actor-7B~\cite{wu2025guiactor} & 60.2 & 54.2 & 58.1 & 30.9 & 0.0 & 49.5\\
Jedi-7B~\cite{xie2025osworldg} & 65.9 & 55.5 & 57.7 & 46.9 & 7.4 & 54.1\\
FocusUI-7B ($r=100\%$)~\cite{ouyang2026focusui} & 63.6 & 61.2 & 63.6 & 34.9 & 0.0 & 54.4\\
UI-TARS-1.5-7B~\cite{qin2025uitars} & 70.1 & 57.9 & 59.7 & 51.7 & 0.0 & 56.0\\
MAI-UI-8B~\cite{zhou2025maiui} & 72.0 & 63.3 & 66.0 & 51.0 & -- & 60.1\\
POINTS-GUI-G-8B~\cite{zhao2026pointsguig} & 73.9 & \underline{73.6} & 70.8 & 55.7 & -- & 66.0\\
UI-Venus-1.5-8B~\cite{gao2026uivenus} & \underline{79.7} & \textbf{76.1} & \underline{72.3} & \underline{60.4} & 22.2 & \underline{69.7}\\
\midrule
\rowcolor{black!7}
\multicolumn{7}{@{}l}{Qwen3-VL-8B configurations}\\
Qwen3-VL-8B-Instruct~\cite{bai2025qwen3vl} & 73.2 & 57.0 & 59.3 & 51.7 & \textbf{53.7} & 57.6\\
GUI-Libra-8B ASFT, thinking~\cite{yang2026guilibra} & 67.4 & 48.8 & 53.8 & 42.3 & 0.0 & 49.3\\
GUI-Libra-8B ASFT, direct~\cite{yang2026guilibra} & 66.3 & 47.0 & 52.6 & 43.6 & 0.0 & 47.7\\
GUI-Libra-8B ASFT+RL, thinking~\cite{yang2026guilibra} & 72.8 & 57.6 & 59.7 & 49.0 & 0.0 & 56.6\\
\textbf{RISE} (Qwen3-VL-8B) & 77.8 & 64.5 & 66.4 & 56.4 & 37.0 & 63.1\\
\midrule
\rowcolor{black!7}
\multicolumn{7}{@{}l}{GUI-Owl-1.5-8B configurations}\\
GUI-Owl-1.5-8B~\cite{xu2026mobileagent35} & 75.1 & 70.9 & 71.9 & 54.4 & \underline{44.4} & 66.3\\
\textbf{RISE} (GUI-Owl-8B) & \textbf{81.2} & \underline{73.6} & \textbf{74.3} & \textbf{63.1} & \underline{44.4} & \textbf{70.6}\\
\bottomrule
\end{tabular*}
\end{table*}

\begin{table}[t]
\centering
\scriptsize
\caption{\textbf{Accuracy (\%) on OSWorld-G-Refined.}
TM: text matching; ER: element recognition; LU: layout understanding; FM: fine-grained manipulation; Ref.: refusal.}
% \caption{\textbf{Accuracy (\%) on OSWorld-G-Refined.}
% TM/ER/LU/FM: text matching, element recognition, layout understanding, and fine-grained manipulation; Ref.: refusal.}
\label{tab:osworldg-refined}
\setlength{\tabcolsep}{2pt}
\renewcommand{\arraystretch}{1.04}
\begin{tabular*}{\columnwidth}{@{\extracolsep{\fill}}lrrrrrr@{}}
\toprule
Method & TM & ER & LU & FM & Ref. & Overall \\
\midrule
UI-TARS-1.5-7B~\cite{qin2025uitars}
& 52.6 & 75.4 & 72.4 & 66.7 & 0.0 & 64.2 \\
GTA1-7B (2507)~\cite{yang2025gta1}
& 63.2 & \textbf{82.1} & 74.2
& \underline{70.5} & 0.0 & 67.7 \\
MAI-UI-8B~\cite{zhou2025maiui}
& 77.4 & 73.0 & 78.3 & 55.7 & -- & 68.6 \\
GUI-Owl-1.5-8B-I~\cite{xu2026mobileagent35}
& 73.1 & 69.4 & 71.0
& \textbf{74.8} & \underline{42.5} & 69.3 \\
UI-Venus-1.5-8B~\cite{gao2026uivenus}
& \underline{82.4} & \underline{81.5} & \textbf{80.2}
& 59.7 & 22.2 & \underline{74.1} \\
\midrule
RISE (GUI-Owl-8B)
& \textbf{86.2} & 78.8 & \underline{78.7}
& 65.8 & \textbf{42.6} & \textbf{75.2} \\
\bottomrule
\end{tabular*}
\end{table}

\section{Experiments}
\label{sec:experiments}
We evaluate RISE on two backbones for GUI grounding and offline action prediction, with matched SFT comparisons, supervision ablations, and inference-cost measurements.

\subsection{Experimental Setup}
\par\noindent\textbf{Benchmarks.}\quad
For GUI grounding, we use OSWorld-G~\cite{xie2025osworldg} with original and refined instructions, ScreenSpot-v2 (SS-v2), UI-I2E-Bench~\cite{liu2025uie2isynth}, and MMBench-GUI L2~\cite{wang2025mmbenchgui}. For offline action prediction, we use the GUI-Libra adaptations~\cite{yang2026guilibra} of AndroidControl~\cite{li2024androidcontrol} and Mind2Web~\cite{deng2023mind2web}, namely AndroidControl-v2 (AC-v2) and MM-Mind2Web-v2, evaluating action types, spatial arguments, and values.

\par\noindent\textbf{Configurations.}\quad
RISE uses eight implicit states on both backbones and preserves their native action formats and image budgets. Both Qwen and GUI-Owl are initially trained on the same 81,418 examples. GUI-Owl undergoes additional training on 167,684 records to further improve grounding performance. Training uses up to eight PPU-ZW810E accelerators (96\,GiB each). Data splits, optimization settings, and evaluation protocols are detailed in Supplementary Material.

\begin{table*}[t]
\centering\footnotesize
\caption{ScreenSpot-v2 accuracy (\%) across platforms and element types. Results are grouped by mobile, desktop, and web interfaces, with text and icon targets reported separately. Overall accuracy is computed over all 1,272 examples. Published references retain their original protocols. Bold and underlined scores indicate the best and second-best results within each group, respectively.}
\label{tab:ssv2-main}
\setlength{\tabcolsep}{2pt}
\renewcommand{\arraystretch}{1.10}
\begin{tabular*}{\textwidth}{@{\extracolsep{\fill}}lrrrrrrr@{}}
\toprule
\multirow{3}{*}{Method / Configuration} & \multicolumn{2}{c}{Mobile} & \multicolumn{2}{c}{Desktop} & \multicolumn{2}{c}{Web} & \multirow{3}{*}{Overall$\uparrow$}\\
\cmidrule(lr){2-3}\cmidrule(lr){4-5}\cmidrule(lr){6-7}
& Text & Icon & Text & Icon & Text & Icon &\\
& $n=290$ & $n=211$ & $n=194$ & $n=140$ & $n=234$ & $n=203$ &\\
\midrule
\rowcolor{black!7}
\multicolumn{8}{@{}l}{External baselines}\\
Aguvis-7B~\cite{xu2024aguvis} & 95.5 & 77.3 & 95.4 & 77.9 & 91.0 & 72.4 & 86.0\\
UGround-V1-7B~\cite{gou2024uground} & 95.0 & 83.3 & 95.0 & 77.8 & 92.1 & 77.2 & 87.6\\
UI-TARS-7B~\cite{qin2025uitars} & 96.9 & 89.1 & 95.4 & 85.0 & 93.6 & 85.2 & 91.6\\
Jedi-7B~\cite{xie2025osworldg} & 96.9 & 87.2 & 95.9 & 87.9 & 94.4 & 84.2 & 91.7\\
GUI-Actor-7B~\cite{wu2025guiactor} & 97.6 & 88.2 & 96.9 & 85.7 & 93.2 & 86.7 & 92.1\\
OpenCUA-7B~\cite{wang2025opencua} & 93.8 & 88.6 & 96.9 & 87.1 & 88.0 & 83.3 & 89.9\\
GTA1-7B~\cite{yang2025gta1} & \underline{99.0} & 88.6 & 94.9 & 89.3 & 92.3 & 86.7 & 92.4\\
FocusUI-7B ($r=100\%$)~\cite{ouyang2026focusui} & 98.8 & 91.6 & 95.6 & \underline{92.1} & 95.0 & 84.4 & 93.1\\
GUI-G$^{2}$-7B~\cite{tang2025guig2} & 98.3 & \underline{91.9} & 95.4 & 89.3 & 94.0 & 87.7 & 93.3\\
InfiGUI-G1-7B (Top-1)~\cite{liu2025infiguig1} & \underline{99.0} & \underline{91.9} & 94.3 & 82.1 & \underline{97.9} & 89.2 & 93.5\\
UI-Venus-Ground-7B~\cite{gu2025uivenus} & \underline{99.0} & 90.0 & 97.0 & 90.7 & 96.2 & 88.7 & 94.1\\
MAI-UI-8B~\cite{zhou2025maiui} & \textbf{99.3} & 88.6 & \underline{98.5} & 90.0 & \underline{97.9} & 90.6 & 94.7\\
POINTS-GUI-G-8B~\cite{zhao2026pointsguig} & \underline{99.0} & 90.5 & \textbf{99.5} & \textbf{92.9} & 94.9 & \underline{92.6} & \underline{95.2}\\
UI-Venus-1.5-8B~\cite{gao2026uivenus} & \textbf{99.3} & \textbf{92.9} & 96.4 & \textbf{92.9} & \textbf{98.3} & \textbf{93.1} & \textbf{95.9}\\
\midrule
\rowcolor{black!7}
\multicolumn{8}{@{}l}{Qwen3-VL-8B configurations}\\
Qwen3-VL-Instruct~\cite{bai2025qwen3vl} & 94.1 & \underline{89.6} & \underline{98.5} & \underline{90.0} & \underline{94.9} & \underline{88.2} & \underline{92.8}\\
GUI-Libra ASFT, thinking~\cite{yang2026guilibra} & 94.8 & 79.6 & 96.4 & 75.7 & 91.9 & 77.3 & 87.1\\
GUI-Libra ASFT, direct~\cite{yang2026guilibra} & 96.2 & 83.4 & 96.9 & 81.4 & 90.6 & 77.3 & 88.5\\
GUI-Libra ASFT+RL, thinking~\cite{yang2026guilibra} & \underline{97.6} & 87.7 & 95.9 & 85.0 & 91.5 & 81.8 & 90.6\\
\textbf{RISE} (Qwen3-VL) & \textbf{99.7} & \textbf{90.0} & \textbf{99.0} & \textbf{92.1} & \textbf{96.2} & \textbf{90.6} & \textbf{95.0}\\
\midrule
\rowcolor{black!7}
\multicolumn{8}{@{}l}{GUI-Owl-1.5-8B configurations}\\
GUI-Owl-1.5~\cite{xu2026mobileagent35} & \underline{98.6} & \underline{90.5} & \underline{97.9} & \underline{87.1} & \underline{95.7} & \underline{87.7} & \underline{93.6}\\
\textbf{RISE} (GUI-Owl) & \textbf{99.7} & \textbf{92.4} & \textbf{98.5} & \textbf{89.3} & \textbf{97.0} & \textbf{89.2} & \textbf{95.0}\\
\bottomrule
\end{tabular*}
\end{table*}

\subsection{Main Results}
\label{sec:osworldg-comparison}
\par\noindent\textbf{OSWorld-G.}\quad
As shown in Table~\ref{tab:osworldg-main}, RISE with GUI-Owl achieves the highest overall accuracy among the listed configurations at 70.6\%, outperforming its backbone by 4.3 percentage points and the strongest published reference, UI-Venus-1.5-8B, by 0.9 points. RISE with Qwen achieves 63.1\% accuracy, improving upon its pretrained backbone by 5.5 points, ASFT with explicit reasoning by 13.8 points, and its ASFT+RL counterpart by 6.5 points. 
% These improvements are achieved without generating intermediate reasoning text.

Across task categories, RISE with GUI-Owl achieves the best results in text matching, layout understanding, and fine-grained manipulation, reaching 81.2\%, 74.3\%, and 63.1\% accuracy, respectively. These results improve upon its backbone by 6.1, 2.4, and 8.7 percentage points, while exceeding the next-best listed methods by 1.5, 2.0, and 2.7 points. For element recognition, RISE achieves 73.6\% accuracy, improving upon its backbone by 2.7 points and tying POINTS-GUI-G for second place, behind UI-Venus-1.5 at 76.1\%. Refusal accuracy remains at 44.4\%, matching its backbone and ranking second jointly, below Qwen3-VL-Instruct at 53.7\%. Overall, RISE improves multiple grounding capabilities, with the largest gain in fine-grained manipulation.

Table~\ref{tab:osworldg-refined} further presents results with refined instructions. RISE achieves 75.2\% accuracy with 424 correct predictions out of 564 examples, compared with 70.6\% and 398 correct predictions under the original instructions. This result exceeds the reported UI-Venus-1.5-8B performance by 1.1 percentage points without additional zooming or verification calls. 
RISE also achieves the highest text-matching (TM) and refusal (Ref.) accuracies of 86.2\% and 42.6\%, surpassing UI-Venus-1.5-8B by 3.8 and 20.4 percentage points, respectively. Across both original and refined instructions, the strong text-matching results suggest that RISE's action-relevant implicit states help capture textual cues for identifying the corresponding GUI elements. RISE also remains competitive in other task categories.

\par\phantomsection\noindent\textbf{ScreenSpot-v2.}\quad
\label{sec:ssv2-comparison}
\label{sec:additional-grounding}
As shown in Table~\ref{tab:ssv2-main}, RISE with Qwen achieves 95.0\% accuracy, outperforming its backbone by 2.3 percentage points and GUI-Libra ASFT and ASFT+RL with explicit reasoning by 7.9 and 4.4 points, respectively. It ranks third among the listed configurations, surpassing MAI-UI-8B at 94.7\%, while trailing POINTS-GUI-G-8B and UI-Venus-1.5-8B at 95.2\% and 95.9\%, respectively. RISE with Qwen achieves the highest accuracy on mobile text at 99.7\%, ranks second on desktop text at 99.0\%, and ties for second on desktop icons at 92.1\%. It improves upon its backbone across all six platform groups, with the largest gain of 5.6 percentage points on mobile text.

RISE with GUI-Owl achieves 95.0\% accuracy, outperforming its backbone by 1.4 percentage points. Improvements are observed across all six groups, with the largest gain on desktop icons, increasing from 87.1\% to 89.3\%.

\par\phantomsection\noindent\textbf{Inference Efficiency.}\quad
\label{sec:deployment-cost}
As shown in Table~\ref{tab:inference-cost}, RISE eliminates intermediate reasoning text while retaining compact action outputs. RISE with Qwen reduces output tokens by 82.1\% compared with explicit-reasoning ASFT+RL and decreases inference time from 15.7 to 3.1 seconds per OSWorld-G example, comparable to the 3.2 seconds of direct Qwen. RISE with GUI-Owl averages 2.2 and 2.6 seconds per example on OSWorld-G and SS-v2, respectively, generating 30.2 and 30.7 output tokens. Compared with action-only SFT, RISE maintains comparable inference latency, with slightly longer outputs mainly due to finer-grained decimal coordinates (e.g., \texttt{[500.2, 130.7]} instead of \texttt{[500, 130]}), rather than additional reasoning tokens.

\newcommand{\risetableinput}[1]{\begingroup\renewenvironment{table}[1][]{}{}\input{#1}\par\endgroup}

\raggedbottom
\makeatletter
\setlength{\@fptop}{0pt}
\setlength{\@dblfptop}{0pt}
\makeatother
\setcounter{topnumber}{1}
\setcounter{dbltopnumber}{2}
\renewcommand{\topfraction}{0.99}
\renewcommand{\textfraction}{0.01}
\renewcommand{\dbltopfraction}{0.99}
\setlength{\textfloatsep}{12pt plus 1pt minus 1pt}

\begin{figure*}[t]
\centering
\captionsetup{hypcap=false,type=figure}
\includegraphics[width=\textwidth]{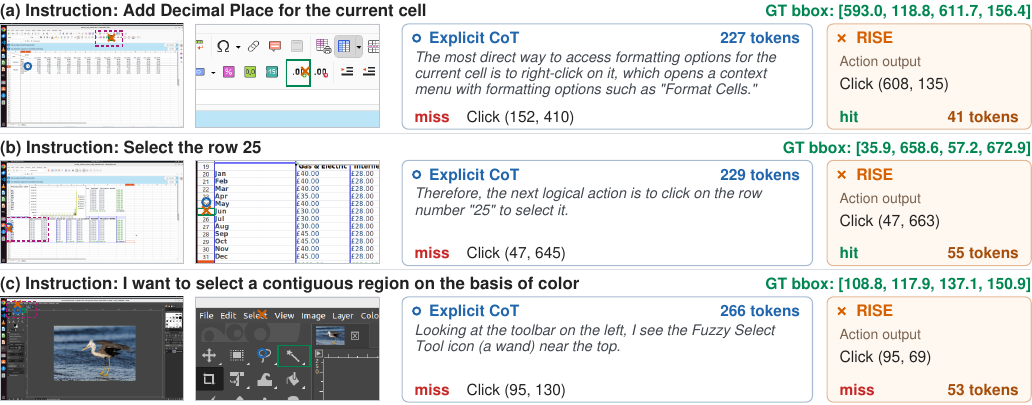}
\caption{\textbf{Explicit CoT versus RISE on OSWorld-G.} Three cases compare GUI-Libra ASFT and RISE. Italic text quotes CoT excerpts; token counts include the complete generated output. Dashed boxes locate enlarged regions. GT boxes use $[x_{\min},y_{\min},x_{\max},y_{\max}]$ on the same 0--1000 scale as predicted clicks. \textcolor[HTML]{008650}{Green} boxes mark target regions; \textcolor[HTML]{2166AC}{blue} circles and \textcolor[HTML]{D66000}{orange} crosses denote CoT and RISE predictions.}
\label{fig:qualitative}
\end{figure*}

\par\phantomsection\noindent\textbf{Offline Action Prediction.}\quad
\label{sec:offline-action}
AndroidControl-v2 (AC-v2) evaluates mobile action prediction under high-level goals and low-level step instructions, while MM-Mind2Web-v2 (MM-v2) evaluates web actions across task, website, and domain splits. As shown in Table~\ref{tab:navigation}, RISE with Qwen achieves 62.3\%, 88.7\%, and 47.1\% accuracy on AC High, AC Low, and MM-v2, respectively. RISE with GUI-Owl achieves 64.8\%, 86.2\%, and 45.5\%.

Compared with action-only SFT, RISE with GUI-Owl improves these three metrics by 5.5, 2.8, and 0.9 percentage points, with the largest gain on high-level goals. RISE with Qwen outperforms the explicit-CoT reference by 5.5, 18.6, and 3.5 points, respectively, showing the largest improvement on low-level instructions. These results suggest that RISE can preserve action-relevant information across different instruction granularities without generating intermediate reasoning text.

\par\phantomsection\noindent\textbf{Transfer Across Benchmarks.}\quad
\label{sec:grounding-transfer}
Table~\ref{tab:i2e-mmbench} reports the performance of RISE with both backbones on UI-I2E-Bench and MMBench-GUI L2. RISE with Qwen improves over its backbone by 2.7 and 1.3 percentage points on the two benchmarks, respectively, while RISE with GUI-Owl achieves gains of 1.5 and 1.4 points. These consistent improvements support the effectiveness of RISE across additional GUI evaluation settings. Prompt selection and evaluation protocols are detailed in Supplementary Material.

\begin{table}[t]
\centering
\footnotesize
\caption{Mean output tokens and seconds per action on the complete OSWorld-G and SS-v2.}
\label{tab:inference-cost}
\setlength{\tabcolsep}{2pt}
\renewcommand{\arraystretch}{1.02}
\begin{tabular*}{\columnwidth}{@{\extracolsep{\fill}}lrrrr@{}}
\toprule
& \multicolumn{2}{c}{OS-G} & \multicolumn{2}{c}{SS-v2}\\
\cmidrule(lr){2-3}\cmidrule(lr){4-5}
Configuration & Tokens & Sec. & Tokens & Sec.\\
\midrule
\rowcolor{black!7}
\multicolumn{5}{@{}l}{\textit{Explicit-reasoning reference}}\\
GUI-Libra ASFT+RL~\cite{yang2026guilibra} & 259.7 & 15.7 & 202.6 & 12.3\\
\midrule
\rowcolor{black!7}
\multicolumn{5}{@{}l}{\textit{Qwen3-VL action-output protocol}}\\
Qwen3-VL, direct~\cite{bai2025qwen3vl} & 47.0 & 3.2 & 45.5 & 3.1\\
\textbf{RISE} (Qwen) & 46.6 & 3.1 & 45.3 & 3.1\\
\midrule
\rowcolor{black!7}
\multicolumn{5}{@{}l}{\textit{GUI-Owl tool-call protocol}}\\
GUI-Owl, original~\cite{xu2026mobileagent35} & 34.0 & 2.4 & 34.7 & 2.8\\
Action-only SFT & 26.0 & 2.0 & 26.7 & 2.4\\
\textbf{RISE} & 30.2 & 2.2 & 30.7 & 2.6\\
\bottomrule
\end{tabular*}
\end{table}

\begin{table}[t]
\centering
\footnotesize
\caption{Offline next-action accuracy (\%) on AC-v2 at High and Low instruction levels and MM-Mind2Web-v2 (macro). Bold and underlined scores indicate the best and second-best results.}
\label{tab:navigation}
\setlength{\tabcolsep}{2pt}
\begin{tabular*}{\columnwidth}{@{\extracolsep{\fill}}lrrr@{}}
\toprule
Configuration & AC High & AC Low & MM macro\\
\midrule
GUI-R1-7B~\cite{luo2025guir1} & 39.7 & 62.3 & 36.9\\
Aguvis-7B~\cite{xu2024aguvis} & 37.7 & 48.0 & 35.4\\
UI-TARS-1.5-7B~\cite{qin2025uitars} & 45.2 & 48.5 & 34.7\\
GLM-4.1V-9B-Thinking~\cite{glmteam2025glm4v} & 37.2 & 67.1 & 26.2\\
Qwen2.5-VL-7B~\cite{bai2025qwen25vl} & 46.5 & 67.8 & 32.5\\
Qwen2.5-VL-32B~\cite{bai2025qwen25vl} & 49.0 & 78.4 & 44.9\\
\midrule
Qwen3-VL-8B~\cite{bai2025qwen3vl} & 54.8 & 77.6 & 43.8\\
GUI-Libra ASFT, thinking (Qwen)~\cite{yang2026guilibra} & 56.8 & 70.1 & 43.6\\
\textbf{RISE (Qwen)} & \underline{62.3} & \textbf{88.7} & \textbf{47.1}\\
\midrule
GUI-Owl-1.5-8B, original~\cite{xu2026mobileagent35} & 15.8 & 53.3 & 14.9\\
Action-only SFT (GUI-Owl) & 59.3 & 83.4 & 44.6\\
\textbf{RISE (GUI-Owl)} & \textbf{64.8} & \underline{86.2} & \underline{45.5}\\
\bottomrule
\end{tabular*}
\end{table}

\begin{table}[t]
\begin{minipage}[t]{\columnwidth}
\vspace{0pt}

\centering\footnotesize
\caption{Grounding accuracy (\%) on UI-I2E-Bench and MMBench-GUI L2. Bold scores indicate the best results, while underlined scores indicate the second-best results in each column.}
\label{tab:gui-matched}
\label{tab:i2e-mmbench}
\setlength{\tabcolsep}{3pt}
\begin{tabular*}{\columnwidth}{@{\extracolsep{\fill}}lrr@{}}
\toprule
Configuration & UI-I2E & \shortstack{MMB-GUI L2}\\
\midrule
Qwen3-VL-Instruct~\cite{bai2025qwen3vl} & \underline{78.5} & 82.9\\
% RISE (Qwen3-VL, + alignment) & \textbf{80.6} & \textbf{83.6}\\
RISE (Qwen3-VL) & \textbf{81.2} & \textbf{84.2}\\
\midrule
GUI-Owl-1.5~\cite{xu2026mobileagent35} & 74.4 & 81.9\\
RISE (GUI-Owl) & 75.9 & \underline{83.3}\\
\bottomrule
\end{tabular*}

\par\vspace{6pt}
\begingroup
\centering\footnotesize

\caption{Supervision-loss ablation of RISE (Qwen) on SS-v2. CE denotes action cross-entropy, and KL denotes action-distribution supervision. Accuracy (\%) is evaluated on all 1,272 examples.}
\label{tab:loss-ablation}
\setlength{\tabcolsep}{1.5pt}
\renewcommand{\arraystretch}{1.08}
\begin{tabular*}{\columnwidth}{@{\extracolsep{\fill}}lccccc@{}}
\toprule
Configuration & \shortstack{CE} & State & \shortstack{KL} & Boundary & SS-v2\\
\midrule
Action supervision only & $\checkmark$ & -- & -- & -- & 93.6\\
% Action supervision only & $\checkmark$ & -- & -- & -- & 90.5\\
w/o State alignment & $\checkmark$ & -- & $\checkmark$ & $\checkmark$ & 94.7\\
w/o Action distribution & $\checkmark$ & $\checkmark$ & -- & $\checkmark$ & 94.2\\
w/o Boundary alignment & $\checkmark$ & $\checkmark$ & $\checkmark$ & -- & \underline{94.8}\\
\midrule
Full RISE & $\checkmark$ & $\checkmark$ & $\checkmark$ & $\checkmark$ & \textbf{95.0}\\
\bottomrule
\end{tabular*}

\endgroup
\end{minipage}
\end{table}

\begin{table}[t]
\begin{minipage}[t]{\columnwidth}
\vspace{0pt}
\captionsetup{hypcap=false,type=table}
\centering\footnotesize

\caption{\textbf{Supervision strategies on Qwen3-VL-8B.}
Accuracy (\%) is reported for each selected configuration.
$K$ denotes the number of implicit states, and Avg. is the equally weighted mean of OSWorld-G (OS-G) and SS-v2.
Bold and underlined scores indicate the best and second-best results, respectively.}
\label{tab:qwen-ablation}
\setlength{\tabcolsep}{2pt}
\renewcommand{\arraystretch}{1.03}
\setlength{\aboverulesep}{1.2pt}
\setlength{\belowrulesep}{1.2pt}
\begin{tabular*}{\columnwidth}{@{\extracolsep{\fill}}lrrrr@{}}
\toprule
Supervision & $K$ & OS-G & SS-v2 & Avg.\\
\midrule
Action-only SFT
& 0 & 58.9 & 91.7 & 75.3\\
Four-field latent supervision
& 8 & 61.3 & 90.5 & 75.9\\
Explicit semantics and action
& 0 & 59.2 & 93.0 & 76.1\\
MIRAGE-style supervision
& 8 & 60.8 & 92.8 & 76.8\\
Full-CoT state alignment
& 8 & 62.8 & 93.2 & 78.0\\
Action-supervised implicit states
& 8 & \underline{63.3} & 93.6 & 78.4\\
Progressive CoT supervision
& 8 & 62.8 & \underline{94.7} & 78.7\\
Operation--target alignment
& 8 & \textbf{63.5} & 94.1 & \underline{78.8}\\
\midrule
RISE (full supervision)
& 8 & 63.1 & \textbf{95.0} & \textbf{79.1}\\
\bottomrule
\end{tabular*}

\par\vspace{6pt}
\begingroup\renewenvironment{table}[1][]{}{}
\begin{table}[t]
\centering\footnotesize
\caption{Latent-state count ablation of RISE with Qwen3-VL-8B. Accuracy (\%) is reported on OSWorld-G and SS-v2, with Avg. denoting their equally weighted mean. Bold and underlined scores indicate the best and second-best results, respectively.}
\label{tab:latent-count}
\setlength{\tabcolsep}{3pt}
\begin{tabular*}{\columnwidth}{@{\extracolsep{\fill}}lrrr@{}}
\toprule Latent states $K$ & OS-G & SS-v2 & Avg.\\\midrule
1 & 62.2 & 94.6 & 78.4\\
4 & \underline{62.9} & \textbf{95.0} & \underline{79.0}\\
8 & \textbf{63.1} & \textbf{95.0} & \textbf{79.1}\\
16 & 62.1 & \underline{94.8} & 78.4\\
\bottomrule
\end{tabular*}
\end{table}

\par\endgroup

\end{minipage}
\end{table}

\subsection{Qualitative Analysis}
\label{sec:qualitative}
Figure~\ref{fig:qualitative} compares explicit CoT and RISE with Qwen on the same screenshots and instructions. In Fig.~\ref{fig:qualitative}(a), CoT selects a cell context menu, whereas RISE correctly identifies the decimal-place toolbar control. In Fig.~\ref{fig:qualitative}(b), CoT identifies row 25 in its reasoning but clicks row 24, while RISE correctly selects row 25. In Fig.~\ref{fig:qualitative}(c), both methods fail to locate the Fuzzy Select Tool, despite CoT identifying it in its reasoning. Across these three examples, RISE generates 41, 55, and 53 output tokens, compared with 227, 229, and 266 for CoT, respectively. These examples illustrate more concise action generation and improved predictions in two cases, while also revealing a shared failure. 

\subsection{Ablation Studies}
\label{sec:alignment-ablation}
\label{sec:gui-matched-ablation}

\par\noindent\textbf{Contribution of Supervision Losses.}\quad
Table~\ref{tab:loss-ablation} evaluates three complementary supervision objectives on SS-v2, with action cross-entropy retained in all configurations. Full RISE achieves 95.0\% accuracy, outperforming action supervision alone by 1.4 percentage points. Removing any of the three objectives reduces performance, with the largest drop of 0.8 points observed when action-distribution supervision is removed. State and boundary alignment further contribute improvements of 0.3 and 0.2 points, respectively. These results demonstrate the complementary benefits of integrating reasoning representations, action distributions, and boundary states for GUI action prediction.

\par\noindent\textbf{Supervision Strategies.}\quad
Table~\ref{tab:qwen-ablation} compares supervision strategies on Qwen.
All configurations use action supervision.
\textit{Action-only SFT} directly predicts actions, while \textit{Explicit semantics and action} additionally supervises textual operation and target descriptions.
In contrast, \textit{Action-supervised implicit states} introduces eight learnable implicit states without explicit semantic supervision.
It achieves a mean accuracy of 78.4\%, outperforming \textit{Action-only SFT} (75.3\%) and \textit{Explicit semantics and action} (76.1\%).
These results demonstrate the benefit of learning implicit states directly from action supervision.

\textit{Operation--target alignment} aligns implicit states with teacher representations of operation and target descriptions, whereas \textit{Full-CoT state alignment} uses representations from the complete reasoning trace. Building on full-CoT state alignment, \textit{RISE (full supervision)} further incorporates action-distribution and reasoning-to-action boundary alignment, achieving the highest mean accuracy of 79.1\%, with 63.1\% on OSWorld-G and 95.0\% on SS-v2. This exceeds the 78.0\% mean of \textit{Full-CoT state alignment}. Although \textit{Operation--target alignment} achieves the highest OSWorld-G accuracy of 63.5\%, its overall mean accuracy is slightly lower at 78.8\%.

We further evaluate three alternative strategies: \textit{Progressive CoT supervision} gradually removes CoT text during training; \textit{MIRAGE-style supervision} refines implicit states using next-screenshot feature supervision; and \textit{Four-field latent supervision} reconstructs operation descriptions, target descriptions, action types and values, and coordinates. None surpasses \textit{RISE (full supervision)} in mean accuracy.

\par\phantomsection\noindent\textbf{Number of Latent States.}\quad
\label{sec:latent-count}
Table~\ref{tab:latent-count} compares one, four, eight, and sixteen latent states under full-CoT joint supervision. Four and eight states achieve mean accuracies of 79.0\% and 79.1\%, respectively, across the two benchmarks, while one and sixteen states both reach 78.4\%. Therefore, we adopt $K=8$, which achieves the highest mean accuracy with a compact latent interface.

\par\noindent\textbf{Matched Backbone Adaptation.}\quad
We further evaluate RISE on GUI-Owl-1.5-8B against action-only SFT under identical training and evaluation conditions, without subsequent continuation. RISE improves OSWorld-G accuracy from 65.2\% to 67.7\% and SS-v2 from 94.6\% to 94.7\%, demonstrating its effectiveness across different backbones.

\section{Conclusion}
In this paper, we propose RISE (Reasoning with Implicit State Embeddings), an implicit reasoning framework for GUI action prediction. RISE constructs contextualized implicit states from learnable embeddings to capture action-relevant reasoning information without generating intermediate reasoning text. We introduce three complementary supervision strategies, including full-CoT state alignment, action-distribution alignment, and reasoning-to-action boundary alignment, to guide implicit state learning for action prediction. RISE achieves 70.6\% accuracy on OSWorld-G with GUI-Owl, outperforming its backbone by 4.3 percentage points, and 95.0\% on ScreenSpot-v2 with Qwen. Experiments across two backbones and six benchmarks demonstrate improvements in GUI grounding and offline action prediction, while inference analyses show reduced decoding overhead compared with explicit reasoning.

{\small\bibliographystyle{ieeenat_fullname}\bibliography{references}}

@inproceedings{xie2024osworld,
  title = {{OSWorld}: Benchmarking Multimodal Agents for Open-Ended Tasks in Real Computer Environments},
  author = {Xie, Tianbao and Zhang, Danyang and Chen, Jixuan and Li, Xiaochuan and Zhao, Siheng and Cao, Ruisheng and Hua, Toh Jing and Cheng, Zhoujun and Shin, Dongchan and Lei, Fangyu and Liu, Yitao and Xu, Yiheng and Zhou, Shuyan and Savarese, Silvio and Xiong, Caiming and Zhong, Victor and Yu, Tao},
  booktitle = {Advances in Neural Information Processing Systems},
  volume = {37},
  year = {2024},
  doi = {10.52202/079017-1650},
  url = {https://proceedings.nips.cc/paper_files/paper/2024/hash/5d413e48f84dc61244b6be550f1cd8f5-Abstract-Datasets_and_Benchmarks_Track.html}
}

@inproceedings{xie2025osworldg,
  title        = {Scaling Computer-Use Grounding via User Interface Decomposition and Synthesis},
  author       = {Xie, Tianbao and Deng, Jiaqi and Li, Xiaochuan and Yang, Junlin and Wu, Haoyuan and Chen, Jixuan and Hu, Wenjing and Wang, Xinyuan and Xu, Yuhui and Wang, Zekun and others},
  booktitle    = {Advances in Neural Information Processing Systems},
  year         = {2025},
  volume      = {38},
  url         = {https://proceedings.neurips.cc/paper_files/paper/2025/hash/22c868099177ee278eb7baccec649f35-Abstract-Datasets_and_Benchmarks_Track.html}
}

@article{liu2026benchmarking,
  title={Benchmarking and Improving GUI Agents in High-Dynamic Environments},
  author={Liu, Enqi and Pan, Liyuan and Gao, Zhi and Yang, Yan and Shi, Chenrui and Liu, Yang and Wu, Jingrong and Li, Qing},
  journal={arXiv preprint arXiv:2604.25380},
  year={2026}
}

@article{bai2025qwen25vl,
  title        = {{Qwen2.5-VL} Technical Report},
  author       = {Bai, Shuai and Chen, Keqin and Liu, Xuejing and Wang, Jialin and Ge, Wenbin and Song, Sibo and Dang, Kai and Wang, Peng and Wang, Shijie and Tang, Jun and others},
  journal      = {arXiv preprint arXiv:2502.13923},
  year         = {2025},
  url          = {https://arxiv.org/abs/2502.13923}
}

@article{glmteam2025glm4v,
  title={Glm-4.5v and glm-4.1v-thinking: Towards versatile multimodal reasoning with scalable reinforcement learning},
  author={Hong, Wenyi and Yu, Wenmeng and Gu, Xiaotao and Wang, Guo and Gan, Guobing and Tang, Haomiao and Cheng, Jiale and Qi, Ji and Ji, Junhui and Pan, Lihang and others},
  journal={arXiv preprint arXiv:2507.01006},
  year={2025}
}

@article{yang2026guilibra,
  title={Gui-libra: Training native gui agents to reason and act with action-aware supervision and partially verifiable rl},
  author={Yang, Rui and Wu, Qianhui and Wang, Zhaoyang and Chen, Hanyang and Yang, Ke and Cheng, Hao and Yao, Huaxiu and Peng, Baolin and Zhang, Huan and Gao, Jianfeng and others},
  journal={arXiv preprint arXiv:2602.22190},
  year={2026}
}

@article{gu2025uivenus,
  title        = {{UI-Venus} Technical Report: Building High-performance {UI} Agents with {RFT}},
  author       = {Gu, Zhangxuan and Zeng, Zhengwen and Xu, Zhenyu and Zhou, Xingran and Shen, Shuheng and Liu, Yunfei and Zhou, Beitong and Meng, Changhua and Xia, Tianyu and Chen, Weizhi and others},
  journal      = {arXiv preprint arXiv:2508.10833},
  year         = {2025}
}

@article{gao2026uivenus,
  title={Ui-venus-1.5 technical report},
  author={{Venus Team} and Gao, Changlong and Gu, Zhangxuan and Liu, Yulin and Qiu, Xinyu and Shen, Shuheng and Wen, Yue and Xia, Tianyu and Xu, Zhenyu and Zeng, Zhengwen and others},
  journal={arXiv preprint arXiv:2602.09082},
  year={2026}
}

@article{zhao2026pointsguig,
  title        = {{POINTS-GUI-G}: {GUI}-Grounding Journey},
  author       = {Zhao, Zhongyin and Liu, Yuan and Liu, Yikun and Wang, Haicheng and Tian, Le and Zhou, Xiao and You, Yangxiu and Yu, Zilin and Yu, Yang and Zhou, Jie},
  journal      = {arXiv preprint arXiv:2602.06391},
  year         = {2026}
}

@inproceedings{hong2023cogagent,
  title = {{CogAgent}: A Visual Language Model for {GUI} Agents},
  author = {Hong, Wenyi and Wang, Weihan and Lv, Qingsong and Xu, Jiazheng and Yu, Wenmeng and Ji, Junhui and Wang, Yan and Wang, Zihan and Dong, Yuxiao and Ding, Ming and Tang, Jie},
  booktitle = {Proceedings of the IEEE/CVF Conference on Computer Vision and Pattern Recognition},
  pages = {14281--14290},
  year = {2024},
  doi = {10.1109/CVPR52733.2024.01354},
  url = {https://openaccess.thecvf.com/content/CVPR2024/html/Hong_CogAgent_A_Visual_Language_Model_for_GUI_Agents_CVPR_2024_paper.html}
}

@inproceedings{cheng2024seeclick,
  title = {{SeeClick}: Harnessing {GUI} Grounding for Advanced Visual {GUI} Agents},
  author = {Cheng, Kanzhi and Sun, Qiushi and Chu, Yougang and Xu, Fangzhi and Li, Yantao and Zhang, Jianbing and Wu, Zhiyong},
  booktitle = {Proceedings of the 62nd Annual Meeting of the Association for Computational Linguistics (Volume 1: Long Papers)},
  pages = {9313--9332},
  year = {2024},
  doi = {10.18653/v1/2024.acl-long.505},
  url = {https://aclanthology.org/2024.acl-long.505/}
}

@inproceedings{wu2024osatlas,
  title        = {{OS-ATLAS}: Foundation Action Model for Generalist {GUI} Agents},
  author       = {Wu, Zhiyong and Wu, Zhenyu and Xu, Fangzhi and Wang, Yian and Sun, Qiushi and Jia, Chengyou and Cheng, Kanzhi and Ding, Zichen and Chen, Liheng and Liang, Paul Pu and Qiao, Yu},
  booktitle    = {International Conference on Learning Representations},
  year         = {2025},
  url         = {https://proceedings.iclr.cc/paper_files/paper/2025/hash/0faa4bc5f522076947a030273629d4fe-Abstract-Conference.html}
}

@inproceedings{gou2024uground,
  title        = {Navigating the Digital World as Humans Do: Universal Visual Grounding for {GUI} Agents},
  author       = {Gou, Boyu and Wang, Ruohan and Zheng, Boyuan and Xie, Yanan and Chang, Cheng and Shu, Yiheng and Sun, Huan and Su, Yu},
  booktitle    = {International Conference on Learning Representations},
  year         = {2025},
  url         = {https://proceedings.iclr.cc/paper_files/paper/2025/hash/4ca0e369689dadb25a5345ba9755ad6f-Abstract-Conference.html}
}

@inproceedings{wu2025guiactor,
  title        = {{GUI-Actor}: Coordinate-Free Visual Grounding for {GUI} Agents},
  author       = {Wu, Qianhui and Cheng, Kanzhi and Yang, Rui and Zhang, Chaoyun and Yang, Jianwei and Jiang, Huiqiang and Mu, Jian and Peng, Baolin and others},
  booktitle    = {Advances in Neural Information Processing Systems},
  year         = {2025},
  volume      = {38},
  url         = {https://proceedings.neurips.cc/paper_files/paper/2025/hash/16130af940e9dabb43c726119bd3b42e-Abstract-Conference.html}
}

@inproceedings{rawles2024androidworld,
  title = {{AndroidWorld}: A Dynamic Benchmarking Environment for Autonomous Agents},
  author = {Rawles, Christopher and Clinckemaillie, Sarah and Chang, Yifan and Waltz, Jonathan and Lau, Gabrielle and Fair, Marybeth and Li, Alice and Bishop, William and Li, Wei and Campbell-Ajala, Folawiyo and Toyama, Daniel and Berry, Robert and Tyamagundlu, Divya and Lillicrap, Timothy and Riva, Oriana},
  booktitle = {International Conference on Learning Representations},
  year = {2025},
  url = {https://proceedings.iclr.cc/paper_files/paper/2025/hash/01a83bc2f2732a58e6aa731e659e7101-Abstract-Conference.html}
}

@article{bai2025qwen3vl,
  title        = {{Qwen3-VL} Technical Report},
  author       = {Bai, Shuai and Cai, Yuxuan and Chen, Ruizhe and Chen, Keqin and Chen, Xionghui and Cheng, Zesen and Deng, Lianghao and Ding, Wei and Gao, Chang and Ge, Chunjiang and others},
  journal      = {arXiv preprint arXiv:2511.21631},
  year         = {2025}
}

@inproceedings{lin2025showui,
  title        = {{ShowUI}: One Vision-Language-Action Model for {GUI} Visual Agent},
  author       = {Lin, Kevin Qinghong and Li, Linjie and Gao, Difei and Yang, Zhengyuan and Wu, Shiwei and Bai, Zechen and Lei, Stan Weixian and Wang, Lijuan and Shou, Mike Zheng},
  booktitle    = {Proceedings of the IEEE/CVF Conference on Computer Vision and Pattern Recognition},
  year         = {2025}
}

@inproceedings{xu2024aguvis,
  title        = {{Aguvis}: Unified Pure Vision Agents for Autonomous {GUI} Interaction},
  author       = {Xu, Yiheng and Wang, Zekun and Wang, Junli and Lu, Dunjie and Xie, Tianbao and Saha, Amrita and Sahoo, Doyen and Yu, Tao and Xiong, Caiming},
  booktitle    = {Proceedings of the 42nd International Conference on Machine Learning},
  year         = {2025},
  pages       = {69772--69805},
  volume      = {267},
  series      = {Proceedings of Machine Learning Research},
  url         = {https://proceedings.mlr.press/v267/xu25ae.html}
}

@article{qin2025uitars,
  title        = {{UI-TARS}: Pioneering Automated {GUI} Interaction with Native Agents},
  author       = {Qin, Yujia and Ye, Yining and Fang, Junjie and Wang, Haoming and Liang, Shihao and Tian, Shizuo and Zhang, Junda and Li, Jiahao and others},
  journal      = {arXiv preprint arXiv:2501.12326},
  year         = {2025}
}

@article{liu2025infiguir1,
  title        = {{InfiGUI-R1}: Advancing Multimodal {GUI} Agents from Reactive Actors to Deliberative Reasoners},
  author       = {Liu, Yuhang and Li, Pengxiang and Xie, Congkai and Hu, Xavier and Han, Xiaotian and Zhang, Shengyu and Yang, Hongxia and Wu, Fei},
  journal      = {arXiv preprint arXiv:2504.14239},
  year         = {2025}
}

@inproceedings{hao2024coconut,
  title = {Training Large Language Models to Reason in a Continuous Latent Space},
  author = {Hao, Shibo and Sukhbaatar, Sainbayar and Su, DiJia and Li, Xian and Hu, Zhiting and Weston, Jason E. and Tian, Yuandong},
  booktitle = {Proceedings of the Conference on Language Modeling},
  year = {2025},
  url = {https://proceedings.colm.cc/paper_files/paper/2025/hash/ed032b08a8822c3635cdcd961012ce60-Abstract-Conference.html}
}

@inproceedings{shen2025codi,
  title        = {{CODI}: Compressing Chain-of-Thought into Continuous Space via Self-Distillation},
  author       = {Shen, Zhenyi and Yan, Hanqi and Zhang, Linhai and Hu, Zhanghao and Du, Yali and He, Yulan},
  booktitle    = {Proceedings of the 2025 Conference on Empirical Methods in Natural Language Processing},
  pages        = {677--693},
  year         = {2025}
}

@inproceedings{shen2025heima,
  title = {Efficient Reasoning with Hidden Thinking},
  author = {Shen, Xuan and Wang, Yizhou and Zhou, Yufa and Shi, Xiangxi and Zhao, Pu and Wang, Yanzhi and Gu, Jiuxiang},
  booktitle = {Proceedings of the 43rd International Conference on Machine Learning},
  series = {Proceedings of Machine Learning Research},
  volume = {306},
  pages = {111225--111240},
  year = {2026},
  url = {https://proceedings.mlr.press/v306/shen26s.html}
}

@inproceedings{wu2025pccot,
  title        = {Parallel Continuous Chain-of-Thought with Jacobi Iteration},
  author       = {Wu, Haoyi and Teng, Zhihao and Tu, Kewei},
  booktitle    = {Proceedings of the 2025 Conference on Empirical Methods in Natural Language Processing},
  pages        = {914--926},
  year         = {2025}
}

@article{ouyang2026focusui,
  title        = {{FocusUI}: Efficient {UI} Grounding via Position-Preserving Visual Token Selection},
  author       = {Ouyang, Mingyu and Lin, Kevin Qinghong and Shou, Mike Zheng and Ng, Hwee Tou},
  journal      = {arXiv preprint arXiv:2601.03928},
  year         = {2026},
  url          = {https://arxiv.org/abs/2601.03928}
}

@article{xu2026mobileagent35,
  title={Mobile-agent-v3.5: Multi-platform fundamental gui agents},
  author={Xu, Haiyang and Zhang, Xi and Liu, Haowei and Wang, Junyang and Zhu, Zhaozai and Zhou, Shengjie and Hu, Xuhao and Gao, Feiyu and Cao, Junjie and Wang, Zihua and others},
  journal={arXiv preprint arXiv:2602.16855},
  year={2026}
}

@article{zhou2025maiui,
  title        = {{MAI-UI} Technical Report: Real-World Centric Foundation {GUI} Agents},
  author       = {Zhou, Hanzhang and Zhang, Xu and Tong, Panrong and Zhang, Jianan and Chen, Liangyu and Kong, Quyu and Cai, Chenglin and Liu, Chen and Wang, Yue and Zhou, Jingren and Hoi, Steven},
  journal      = {arXiv preprint arXiv:2512.22047},
  year         = {2025},
  url          = {https://arxiv.org/abs/2512.22047}
}

@article{wang2025opencua,
  title={Opencua: Open foundations for computer-use agents},
  author={Wang, Xinyuan and Wang, Bowen and Lu, Dunjie and Yang, Junlin and Xie, Tianbao and Wang, Junli and Deng, Jiaqi and Guo, Xiaole and Xu, Yiheng and Wu, Chen and others},
  journal={Advances in Neural Information Processing Systems},
  volume={38},
  pages={139756--139806},
  year={2026}
}

@inproceedings{yang2025gta1,
  title={Gta1: Gui test-time scaling agent},
  author={Yang, Yan and Li, Dongxu and Dai, Yutong and Yang, Yuhao and Luo, Ziyang and Zhao, Zirui and Hu, Zhiyuan and Huang, Junzhe and Saha, Amrita and Chen, Zeyuan and others},
  booktitle={International Conference on Learning Representations},
  volume={2026},
  pages={48295--48323},
  year={2026}
}

@inproceedings{tang2025guig2,
  title={GUI-G$^2$: Gaussian Reward Modeling for GUI Grounding},
  author={Tang, Fei and Gu, Zhangxuan and Lu, Zhengxi and Liu, Xuyang and Shen, Shuheng and Meng, Changhua and Wang, Wen and Zhang, Wenqi and Shen, Yongliang and Lu, Weiming and others},
  booktitle={Proceedings of the AAAI Conference on Artificial Intelligence},
  volume={40},
  number={39},
  pages={33214--33222},
  year={2026}
}

@inproceedings{liu2025infiguig1,
  title={Infigui-g1: Advancing gui grounding with adaptive exploration policy optimization},
  author={Liu, Yuhang and Liu, Zeyu and Zhu, Shuanghe and Li, Pengxiang and Xie, Congkai and Wang, Jiasheng and Hu, Xueyu and Han, Xiaotian and Yuan, Jianbo and Wang, Xinyao and others},
  booktitle={Proceedings of the AAAI Conference on Artificial Intelligence},
  volume={40},
  number={38},
  pages={32267--32275},
  year={2026}
}

@inproceedings{goyal2024pause,
  title={Think before you speak: Training Language Models With Pause Tokens},
  author={Goyal, Sachin and Ji, Ziwei and Rawat, Ankit Singh and Menon, Aditya Krishna and Kumar, Sanjiv and Nagarajan, Vaishnavh},
  booktitle={International Conference on Learning Representations},
  year={2024},
  url={https://arxiv.org/abs/2310.02226}
}

@article{yang2026mirage,
  title={{MIRAGE}: Mobile Agents with Implicit Reasoning and Generative World Models},
  author={Yang, Zhichao and Hu, Yuanze and Hao, Haojie and Hao, Longkun and Huang, Dongshuo and Lin, Hongyu and Li, Gen and Hong, Lanqing and Lou, Yihang and Bai, Yan},
  journal={arXiv preprint arXiv:2606.04627},
  year={2026},
  url={https://arxiv.org/abs/2606.04627}
}

@inproceedings{liu2025uie2isynth,
  title={{UI-E2I-Synth}: Advancing {GUI} Grounding with Large-Scale Instruction Synthesis},
  author={Liu, Xinyi and Zhang, Xiaoyi and Zhang, Ziyun and Lu, Yan},
  booktitle={Findings of the Association for Computational Linguistics: ACL 2025},
  pages={15668--15684},
  year={2025},
  doi={10.18653/v1/2025.findings-acl.809},
  url={https://aclanthology.org/2025.findings-acl.809/}
}

@article{wang2025mmbenchgui,
  title={Mmbench-gui: Hierarchical multi-platform evaluation framework for gui agents},
  author={Wang, Xuehui and Wu, Zhenyu and Xie, JingJing and Ding, Zichen and Yang, Bowen and Li, Zehao and Liu, Zhaoyang and Li, Qingyun and Dong, Xuan and Chen, Zhe and others},
  journal={arXiv preprint arXiv:2507.19478},
  year={2025}
}

@inproceedings{li2021prefixtuning,
  title={{Prefix-Tuning}: Optimizing Continuous Prompts for Generation},
  author={Li, Xiang Lisa and Liang, Percy},
  booktitle={Proceedings of the 59th Annual Meeting of the Association for Computational Linguistics and the 11th International Joint Conference on Natural Language Processing (Volume 1: Long Papers)},
  pages={4582--4597},
  year={2021},
  doi={10.18653/v1/2021.acl-long.353},
  url={https://aclanthology.org/2021.acl-long.353/}
}

@inproceedings{lester2021prompttuning,
  title={The Power of Scale for Parameter-Efficient Prompt Tuning},
  author={Lester, Brian and Al-Rfou, Rami and Constant, Noah},
  booktitle={Proceedings of the 2021 Conference on Empirical Methods in Natural Language Processing},
  pages={3045--3059},
  year={2021},
  doi={10.18653/v1/2021.emnlp-main.243},
  url={https://aclanthology.org/2021.emnlp-main.243/}
}

@article{deng2023implicitcot,
  title={Implicit Chain of Thought Reasoning via Knowledge Distillation},
  author={Deng, Yuntian and Prasad, Kiran and Fernandez, Roland and Smolensky, Paul and Chaudhary, Vishrav and Shieber, Stuart},
  journal={arXiv preprint arXiv:2311.01460},
  year={2023},
  url={https://arxiv.org/abs/2311.01460}
}

@inproceedings{li2024androidcontrol,
  title={On the Effects of Data Scale on {UI} Control Agents},
  author={Li, Wei and Bishop, William and Li, Alice and Rawles, Chris and Campbell-Ajala, Folawiyo and Tyamagundlu, Divya and Riva, Oriana},
  booktitle={Advances in Neural Information Processing Systems},
  volume={37},
  year={2024},
  url={https://proceedings.neurips.cc/paper_files/paper/2024/hash/a79f3ef3b445fd4659f44648f7ea8ffd-Abstract-Datasets_and_Benchmarks_Track.html}
}

@inproceedings{deng2023mind2web,
  title={{Mind2Web}: Towards a Generalist Agent for the Web},
  author={Deng, Xiang and Gu, Yu and Zheng, Boyuan and Chen, Shijie and Stevens, Samuel and Wang, Boshi and Sun, Huan and Su, Yu},
  booktitle={Advances in Neural Information Processing Systems},
  volume={36},
  year={2023},
  url={https://proceedings.neurips.cc/paper_files/paper/2023/hash/5950bf290a1570ea401bf98882128160-Abstract-Datasets_and_Benchmarks.html}
}

@article{hinton2015distilling,
  title = {Distilling the Knowledge in a Neural Network},
  author = {Hinton, Geoffrey and Vinyals, Oriol and Dean, Jeff},
  journal = {arXiv preprint arXiv:1503.02531},
  year = {2015},
  url = {https://arxiv.org/abs/1503.02531}
}

@inproceedings{lu2025uir1,
  title = {{UI-R1}: Enhancing Efficient Action Prediction of {GUI} Agents by Reinforcement Learning},
  author = {Lu, Zhengxi and Chai, Yuxiang and Guo, Yaxuan and Yin, Xi and Liu, Liang and Wang, Hao and Xiao, Han and Ren, Shuai and Zhao, Pengxiang and Liu, Guangyi and Xiong, Guanjing and Li, Hongsheng},
  booktitle = {Proceedings of the AAAI Conference on Artificial Intelligence},
  volume = {40},
  number = {21},
  pages = {17608--17616},
  year = {2026},
  doi = {10.1609/aaai.v40i21.38816},
  url = {https://ojs.aaai.org/index.php/AAAI/article/view/38816}
}

@article{luo2025guir1,
  title={Gui-r1: A generalist r1-style vision-language action model for gui agents},
  author={Luo, Run and Wang, Lu and He, Wanwei and Chen, Longze and Li, Jiaming and Xia, Xiaobo},
  journal={arXiv preprint arXiv:2504.10458},
  year={2025}
}
\end{document}